\documentclass[12pt]{article}

\newcommand{\BEAS}{\begin{eqnarray*}}
\newcommand{\EEAS}{\end{eqnarray*}}
\newcommand{\BEA}{\begin{eqnarray}}
\newcommand{\EEA}{\end{eqnarray}}
\newcommand{\BEQ}{\begin{equation}}
\newcommand{\EEQ}{\end{equation}}
\newcommand{\BIT}{\begin{itemize}}
\newcommand{\EIT}{\end{itemize}}
\newcommand{\BNUM}{\begin{enumerate}}
\newcommand{\ENUM}{\end{enumerate}}

\newcommand{\BA}{\begin{array}}
\newcommand{\EA}{\end{array}}

\newcommand{\ones}{\mathbf 1}

\newcommand{\reals}{{\mbox{\bf R}}}

\newcommand{\symm}{{\mbox{\bf S}}}  

\newcommand{\Tr}{\mathop{\bf Tr}}

\newcommand{\QED}{~~\rule[-1pt]{6pt}{6pt}}

\newtheorem{remark}[theorem]{Remark}

\newcounter{exno}

{\begin{quote}}{\end{quote}}

\makeatletter
\long\def\@makecaption#1#2{
   \vskip 9pt 
   \begin{small}
   \setbox\@tempboxa\hbox{{\bf #1:} #2}
   \ifdim \wd\@tempboxa > 5.5in
        \begin{center}
        \begin{minipage}[t]{5.5in}
        \addtolength{\baselineskip}{-0.95pt}
        {\bf #1:} #2 \par
        \addtolength{\baselineskip}{0.95pt}
        \end{minipage}
        \end{center}
   \else 
    \hbox to\hsize{\hfil\box\@tempboxa\hfil}  
   \fi
   \end{small}\par
}
\makeatother

\newcounter{oursection}

\newcounter{lecture}

\usepackage{fullpage,times,amsmath}
\usepackage{amsfonts,amssymb}
\usepackage{psfrag,graphics,epsfig,url}

\newcommand{\Card}{\mathop{\bf Card}}

\begin{document}

\title{Clustering and Feature Selection using\\ Sparse Principal Component Analysis}

\author{Ronny Luss\thanks{ORFE Department, Princeton University,
Princeton, NJ 08544. \texttt{rluss@princeton.edu}} \and Alexandre
d'Aspremont\thanks{ORFE Department, Princeton University, Princeton,
NJ 08544. \texttt{aspremon@princeton.edu}}}

\maketitle

\begin{abstract}
In this paper, we study the application of sparse principal component analysis (PCA) to clustering and feature selection problems. Sparse PCA seeks sparse factors, or linear combinations of the data variables, explaining a maximum amount of variance in the data while having only a limited number of nonzero coefficients. PCA is often used as a simple clustering technique and sparse factors allow us here to interpret the clusters in terms of a reduced set of variables. We begin with a brief introduction and motivation on sparse PCA and detail our implementation of the
algorithm in d'Aspremont et al. (2005). We then apply these results to some classic clustering and feature selection problems arising in biology.
\end{abstract}

\\

\section{Introduction}
This paper focuses on applications of sparse principal component analysis to clustering and feature selection problems, with a particular focus on gene expression data analysis. Sparse methods have had a significant impact in many areas of statistics, in particular regression and classification (see \cite{Cand05}, \cite{Dono05} and \cite{Vapn95} among others). As in these areas, our motivation for developing sparse multivariate visualization tools is the potential of these methods for yielding statistical results that are both more interpretable and more robust than classical analyses, while giving up little statistical efficiency.

Principal component analysis (PCA) is a classic tool for analyzing large scale multivariate data. It seeks linear combinations of the data variables (often called factors or principal components) that capture a maximum amount of variance. Numerically, PCA only amounts to computing a few leading eigenvectors of the data's covariance matrix, so it can be applied to very large scale data sets. One of the key shortcomings of PCA however is that these factors are linear combinations of \emph{all} variables; that is, all factor coefficients (or loadings) are non-zero.  This means that while PCA facilitates model interpretation and visualization by concentrating the information in a few key factors, the factors themselves are still constructed using {\em all} observed variables. In many applications of PCA, the coordinate axes have a direct physical interpretation; in finance or biology for example, each axis might correspond to a specific financial asset or gene. In such cases, having only a few nonzero coefficients in the principal components would greatly improve the relevance and interpretability of the factors. In sparse PCA, we seek a trade-off between the two goals of {\em expressive power} (explaining most of the variance or information in the data) and {\em interpretability} (making sure that the factors involve only a few coordinate axes or variables). When PCA is used as a clustering tool, sparse factors will allow us to identify the clusters with the action of only a few variables.

Earlier methods to produce sparse factors include Cadima and Jolliffe \cite{cadi95} where the loadings with smallest absolute value are thresholded to zero and nonconvex algorithms called SCoTLASS by \cite{Joll03}, SLRA \cite{Zhan02a,Zhan04} and SPCA by \cite{Zou04}. This last method works by writing PCA as a regression-type optimization problem and applies LASSO \cite{tibs96}, a penalization technique based on the $l_1$ norm. Very recently, \cite{Mogh06b} and \cite{Mogh06a} also proposed a greedy approach which seeks globally optimal solutions on small problems and uses a greedy method to approximate the solution of larger ones.  In what follows, we give a brief introduction to the relaxation of this problem in \cite{dasp04a} and describe how this smooth optimization algorithm was implemented. The most expensive numerical step in this algorithm is the computation of the gradient as a matrix exponential and our key numerical contribution here is to show that using only a partial eigenvalue decomposition of the current iterate can produce a sufficiently precise gradient approximation while drastically improving computational efficiency. We then show on classic gene expression data sets that using sparse PCA as a simple clustering tool isolates very relevant genes compared to other techniques such as recursive feature elimination or ranking.

The paper is organized as follows. In Section \ref{s:algo}, we begin with a brief introduction and motivation on sparse PCA and detail our implementation of the algorithm in a numerical toolbox called DSPCA, which is available for download on the authors' websites. In Section \ref{s:application}, we describe the application of sparse PCA to clustering and feature selection on gene expression data. 

\section{Algorithm and implementation}
\label{s:algo} In this section, we begin by introducing the sparse PCA problem and the corresponding semidefinite relaxation derived in \cite{dasp04a}. We then discuss how to use a partial eigenvalue decomposition of the current iterate to produce a sufficiently precise gradient approximation and improve computational efficiency.

\subsection{Sparse PCA \& Semidefinite Relaxation}
Given a covariance matrix $A\in\symm_n$, where $\symm_n$ is the set of symmetric matrices of dimension $n$, the problem of finding a sparse factor which explains a maximum amount of variance in the data can be written:
\BEQ \label{eq:variat-prog}
\BA{ll}
\mbox{maximize} & x^TAx\\
\mbox{subject to} & \|x\|_2=1\\
& \Card(x) \leq k,
\EA \EEQ
in the variable $x\in\reals^n$ where $\Card (x)$ denotes the cardinality of $x$ and $k>0$ is a parameter controlling this cardinality. Computing sparse factors with maximum variance is a combinatorial problem and is numerically hard and \cite{dasp04a} use semidefinite relaxation techniques to compute approximate solutions efficiently by solving the following convex program:
\BEQ \label{eq:variat-relax} \BA{lll}
\mbox{maximize} & \Tr(AX)\\
\mbox{subject to} & \Tr(X)=1\\
& \ones ^T |X| \ones \leq k\\
& X \succeq 0,
\EA \EEQ
which is a semidefinite program in the variable $X\in\symm^n$, where $\ones ^T |X| \ones=\sum_{ij} |X_{ij}|$ can be seen as a convex lower bound on the function $\Card(X)$. When the solution of the above problem has rank one, we have $X=xx^T$ where $x$ is an approximate solution to (\ref{eq:variat-prog}). When the solution of this relaxation is not rank one, we use the leading eigenvector of $X$ as a principal component.

While small instances of problem (\ref{eq:variat-relax}) can be solved efficiently using interior point semidefinite programming solvers such as SEDUMI by \cite{stur99}, the $O(n^2)$ linear
constraints make these solvers inefficient for reasonably large
instances. Furthermore, interior point methods are geared towards
solving small problems with high precision requirements, while here we need to solve very large instances with relatively low precision. In \cite{dasp04a} it was shown that a smoothing
technique due to \cite{Nest03} could be applied to problem
(\ref{eq:variat-relax}) to get a complexity estimate of $
O(n^4\sqrt{\log n} / \epsilon)$ and a much lower memory cost per iteration. The key numerical step in this algorithm is the computation of a smooth approximation of problem (\ref{eq:variat-relax}) and the gradient of the objective, which amounts to computing a matrix exponential.

\subsection{Implementation}
\label{s:num-imp} Again, given a covariance matrix $A\in\symm_n$, the DSPCA code solves a penalized formulation of problem
(\ref{eq:variat-relax}), written as:
\BEQ\label{eq:orig-pb} \BA{ll}
\mbox{maximize} & \Tr(AX) - \rho \ones^T |X| \ones\\
\mbox{subject to} & \Tr(X)=1\\
& X \succeq 0,
\EA
\EEQ
in the variable $X\in\symm_n$.  The dual of this program can be written as:
\BEQ\label{eq:orig-dual}
\BA{ll}
\mbox{minimize} & f(U)=\lambda^\mathrm{max}(A+U)\\
\mbox{subject to} & |U_{ij}|\leq \rho.\\
\EA
\EEQ
in the variable $U\in\symm^n$. The algorithm in \cite{dasp04a} regularizes the objective $f(U)$ in (\ref{eq:orig-dual}),
replacing it by the smooth (i.e. with Lipschitz continuous gradient) uniform approximation:
\[
f_\mu(U) = \mu \log \left(\Tr \exp((A+U)/\mu)\right) - \mu \log n,
\]
whose gradient can be computed explicitly as:
\[
\nabla f_\mu(U):=  \exp\left((A+U)/\mu \right)/\left(\Tr \exp\left((A+U)/\mu \right)\right).
\]
Following \cite{nest83}, solving the smooth problem:
\[
\min_{\{U\in{S^n}, |U_{ij}|\leq \rho\}} f_\mu(U)
\]
with $\mu =\epsilon/2\log(n)$ then produces an $\epsilon$-approximate solution to (\ref{eq:orig-pb}) and requires 
\BEQ\label{eq:complexity}
O\left(\rho \frac{n\sqrt{\log n}}{\epsilon}\right)
\EEQ
iterations. The main step at each iteration is computing the matrix exponential
$\exp((A+U)/\mu)$. This is a classic problem in linear algebra (see \cite{Mole03} for a comprehensive survey) and in what follows, we detail three different methods implemented in DSPCA and their relative performance.

\paragraph{Full eigenvalue decomposition}
An exact computation of the matrix exponential can be done
through a full eigenvalue decomposition of the matrix.  Given the spectral decomposition $VDV^T$ of a matrix $A$ where the columns of $V$ are the eigenvectors and $D$ is a diagonal matrix comprised of the eigenvalues $(d_i)_{i=1}^n$, the matrix exponential can be computed as:
\[
\exp(A)=VHV^T,
\]
where $H$ is the diagonal matrix with $(h_i=e^{d_i})_{i=1}^n$ on its diagonal. While this is a simple procedure, it is also relatively inefficient.

\paragraph{Pad\'e approximation}
The next method implemented in DSPCA is called Pad\'e approximation and approximates the exponential by a rational function.  The (p,q) Pad\'e approximation for $\exp(A)$ is defined by (see \cite{Mole03}):
\BEQ\label{eq:pade}
R_{pq}(A)=[D_{pq}]^{-1}N_{pq}(A),
\EEQ
where
\[
N_{pq}(A)=\sum_{j=0}^p \frac{(p+q-j)!p!}{(p+q)!j!(p-j)!}A^j \quad \mbox{and} \quad D_{pq}(A)=\sum_{j=0}^q \frac{(p+q-j)!q!}{(p+q)!j!(q-j)!}(-A)^j.
\]
Here $p$ and $q$ control the degree and precision of the approximation and we set $p=q=6$ (we set $p=q$ in practice due to computational issues; see \cite{Mole03}).  The approximation is only valid in a small neighborhood of zero, which means that we need to scale down the matrix before approximating its exponential using (\ref{eq:pade}), and then scale it back to its original size. This scaling and squaring can be done efficiently using the property that $e^A=(e^\frac{A}{m})^m.$ We first scale the matrix $A$ so that $\frac{1}{m}\|A\|\leq 10^{-6}$ and find the smallest integer $s$ such that this is true for $m=2^s$.  We then use the Pad\'e approximation to compute $e^{A}$, and simply square the result $s$ times to scale it back.

Pad\'e approximation only requires computing one matrix inversion and several matrix products which can be done very efficiently. However, if $n$ or $s$ get somewhat large, scaling and squaring can be costly, in which case a full eigenvalue decomposition has better performance.  While Pad\'e approximations appear to be the current method of choice for computing matrix exponentials (see \cite{Mole03}), it does not perform as well as expected on our problems compared to partial eigenvalue decomposition discussed below, because of the particular structure of our optimization problem. Numerical results illustrating this issue are detailed in the last section.

\paragraph{Partial eigenvalue decomposition}
The first two classic methods we described for computing the exponential of a matrix are both geared towards producing a solution up to (or close to) machine precision. In most of the sparse PCA problems we solve here, the target precision for the algorithm is of the order $10^{-1}$. Computing the gradient up to machine precision in this context is unnecessarily costly. In fact, \cite{dasp05} shows that the optimal convergence of the algorithm in \cite{Nest03} can be achieved using approximate gradient values $\tilde \nabla f_\mu(U)$, provided the error satisfies the following condition:
\BEQ\label{eq:grad-cond}
|\langle \tilde
\nabla f_\mu(U)-\nabla f_\mu(U),Y\rangle| \leq \delta, \quad |U_{ij}|,|Y_{ij}|\leq \rho,~i,j=1,\ldots,n,
\EEQ
where $\delta$ is a parameter controlling the approximation error. In practice, this means that we only need to compute a few leading eigenvalues of the matrix $(A+U)/\mu$ to get a sufficient gradient approximation.  More precisely, if we denote by $\lambda\in\reals^n$ the eigenvalues of $(A+U)/\mu$, condition~(\ref{eq:grad-cond}) can be used to show that, to get convergence, we need only compute the $j$ largest eigenvalues with $j$ such that:
\BEQ \label{DSPCAgrad-cond}
\frac{(n-j)e^{\lambda_j}\sqrt{\sum_{i=1}^j{e^{2\lambda_i}}}}
{(\sum_{i=1}^j{e^{\lambda_i}})^2}+\frac{\sqrt{n-j}e^{\lambda_j}}
{\sum_{i=1}^j{e^{\lambda_i}}}
 \leq \frac{\delta}{\rho n}.
\EEQ
The terms on the left side decrease as $j$ increases meaning the condition is satisfied by increasing the number of eigenvalues used.  Computing the $j$ largest eigenvalues of a matrix can be done very efficiently using packages such as ARPACK if $j<<n$. When $j$ becomes large, the algorithm switches to full eigenvalue decomposition. Finally, the leading eigenvalues tend to coalesce close to the optimum, potentially increasing the number of eigenvalues required at each iteration (see \cite{Pata98} for example) but this phenomenon does not seem to appear at the low precision targets considered here. We detail empirical results on the performance of this technique in the following sections illustrating how this technique clearly dominates the two others for large scale problems.

\subsection{Comparison with interior point solvers}
We give an example illustrating the necessity of a large-scale solver using a smooth approximation to problem (\ref{eq:variat-relax}).  The Sedumi implementation to problem (\ref{eq:variat-relax}) runs out of memory for problem sizes larger than 60, so we compare the example on very small dimensions against our implementation with partial eigenvalue decomposition (DSPCA).  The covariance matrix is formed using colon cancer gene expression data detailed in the following section.  Table \ref{tab:sedumi_partial} shows running times for DSPCA and Sedumi on for various (small) problem dimensions. DSPCA clearly beats the interior point solver in computational time while achieving comparable precision (measured as the percentage of variance explained by the sparse factor). For reference, we show how much variation is explained by the leading principal component. The decrease in variance using Sedumi and DSPCA represents  the cost of sparsity here.

\begin{table}[htp]
\begin{center}
\begin{tabular}{c|cccccc}
Dim. & NonZeros & DSPCA time & DSPCA Var. & Sedumi time& Sedumi Var. & PCA Var. \\ \hline
10 & 3 & 0.08 & 42.88 \% &   0.39 & 43.00 \% & 49.36 \% \\ 
20 & 8 & 0.27 & 45.67 \% &   5.29 & 47.50 \% & 52.58 \% \\ 
30 & 11 & 0.99 & 40.74 \% &  41.99 & 42.44 \% & 51.13 \% \\ 
40 & 14 & 4.55 & 40.26 \% & 229.44 & 41.89 \% & 52.40 \% \\ 
50 & 15 & 7.52 & 36.76 \% & 903.36 & 38.53 \% & 50.24 \% \\ 
\end{tabular}
\caption{CPU time (in seconds) and explained variance for Sedumi and smooth optimization with partial eigenvalue decompositions (DSPCA). Var measures the percentage of total variance explained by the sparse factors. DSPCA significantly outperforms in terms of CPU time while reaching comparable variance targets.}
\label{tab:sedumi_partial}
\end{center}\end{table}

\section{Clustering and Feature Selection}
\label{s:application} In this section, we use our code for sparse PCA (DSPCA), to analyze large sets of gene expression data and we discuss applications of this technique to clustering and feature selection. PCA is very often used as a simple tool for data visualization and clustering (see \cite{Sreb06} for a recent analysis), here sparse factors allow us to interpret the low dimensional projection of the data in terms of only a few variables.

\subsection{Gene expression data} We first test the performance of DSPCA on covariance matrices generated from a gene expression data set from \cite{Alon99} on colon cancer.  The data set is comprised of 62 tissue samples, 22 from normal colon tissues and 40 from cancerous colon tissues with 2000 genes measured in each sample. We preprocess the data by taking the log of all data intensities and then normalize the log data of each sample to be mean zero and standard deviation one, which is a classic procedure to minimize experimental effects (see \cite{Huan05a}).

\begin{figure} [htp]
\begin{tabular*}{\textwidth}{@{\extracolsep{\fill}}cc}
\psfrag{time}[b][t]{\small{CPU time (seconds)}}
\psfrag{dim}[t][b]{\small{Dimension}}
\includegraphics[width=.50 \textwidth]{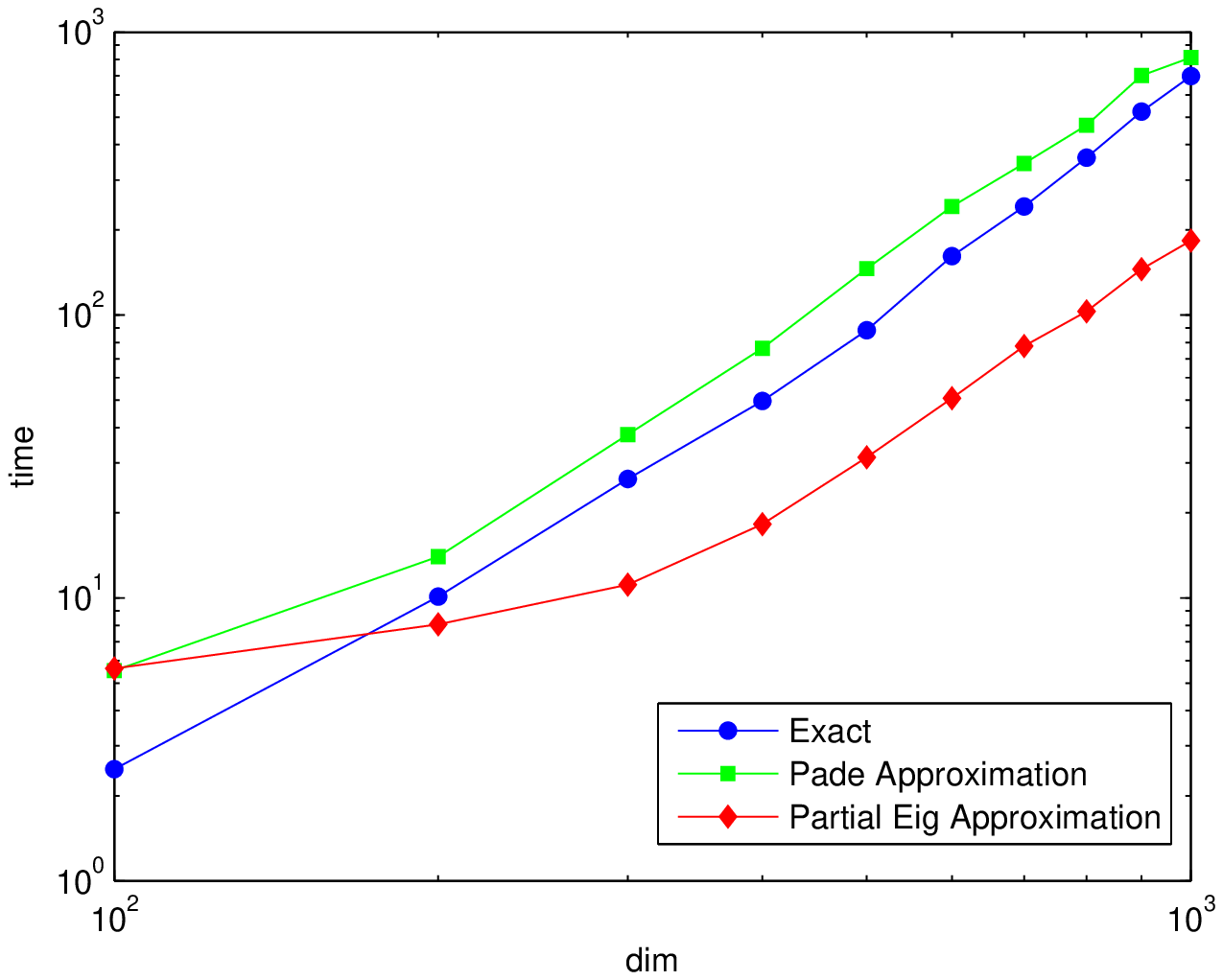} &
\psfrag{eigs}[b][t]{\small{\% eigs. required}} \psfrag{dim}[t][b]{\small{Dimension}}
\includegraphics[width=.50 \textwidth]{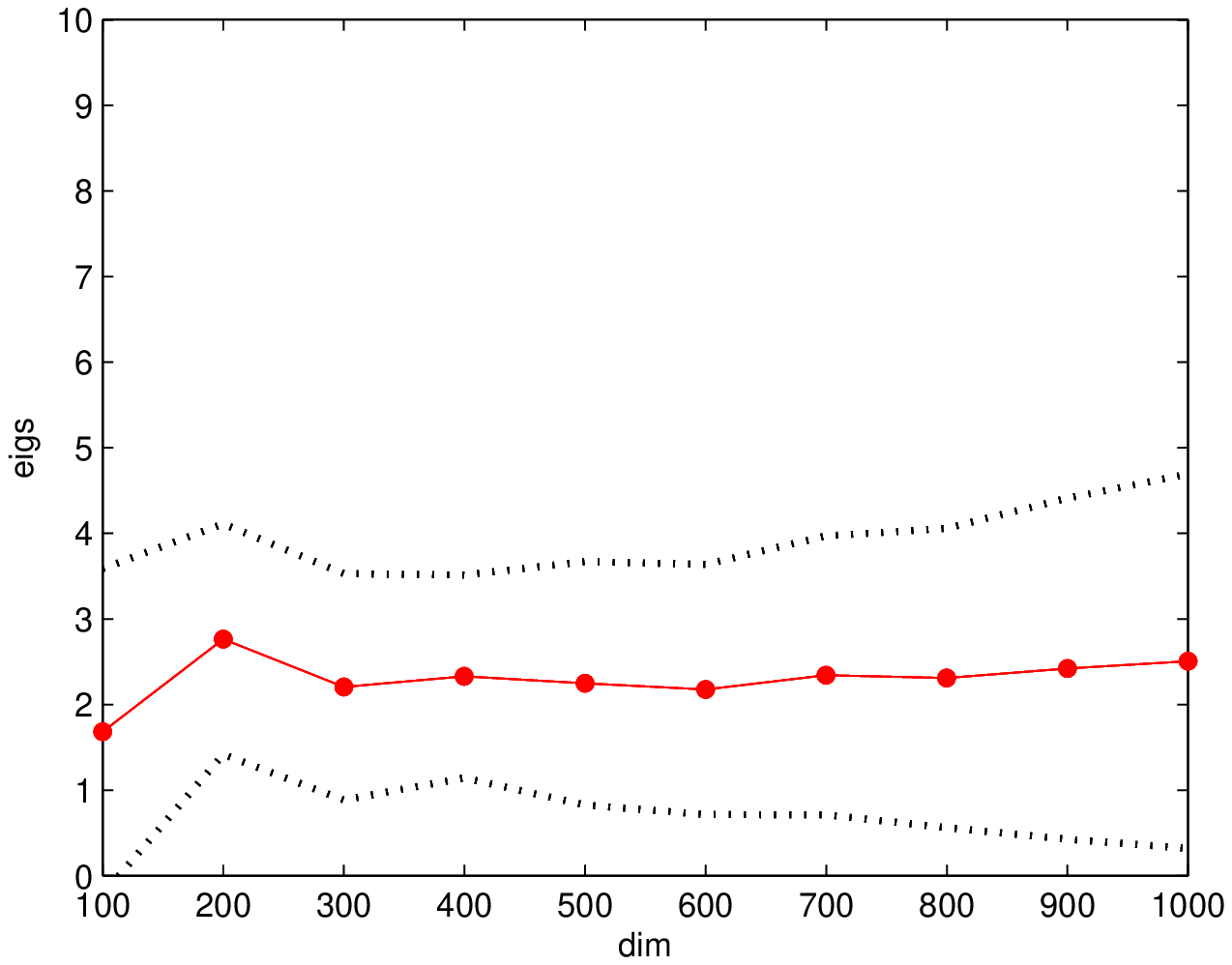}
\end{tabular*}
\caption{Performance of smooth optimization using exact gradient, Pad\'e approximations and partial eigenvalue decomposition. CPU time versus problem dimension (left). Average percentage of eigenvalues required versus problem dimension (right) for partial eigenvalue decomposition, with dashed lines at plus and minus one standard deviation.}
\label{fig:dimension_coloncancer}
\end{figure}

Since the semidefinite programs produced by these data sets are relatively large (with $n=1000$ after preprocessing), we first test the numerical performance of the algorithm and run the code on increasingly large problems using each of the three methods described in section \ref{s:num-imp} (full eigenvalue decomposition, Pad\'e approximation and partial eigenvalue decomposition). Theoretically, the performance increase of using partial, rather than full, eigenvalue decompositions should be substantial when only a few eigenvalues are required.  In practice there is overhead due to the necessity of testing condition (\ref{DSPCAgrad-cond}) iteratively. Figure \ref{fig:dimension_coloncancer} depicts the results of these tests on a 3.0 GHz CPU in a loglog plot of runtime (in seconds) versus problem dimension (on the left). We plot the average number of eigenvalues required by condition (\ref{DSPCAgrad-cond}) versus problem dimension (on the right), with dashed lines at plus and minus one standard deviation.  We cannot include interior point algorithms in this comparison because memory problems occur for dimensions greater than 50.

In these tests, partial eigenvalue is more than three times faster than the other methods for large problem sizes, meaning the approximate gradient is the dominating alternative compared to the exact gradient. The plot of the average percentage of eigenvalues for each test shows that on average only 2-3\% of the eigenvalues are necessary. Given the gain performance, the main limitation becomes memory; the partial eigenvalue implementation has memory allocation problems at dimensions 1900 and larger on a computer with 3.62 Gb RAM. Notice also that in our experiments, computing the matrix exponential using Pad\'e approximation is slower than performing a full eigenvalue decomposition, which reflects the high numerical cost of matrix multiplications necessary for scaling and squaring in Pad\'e approximations.

Next, we test the impact of the sparsity target on computational complexity. We observe in~(\ref{eq:complexity}) that the number of iterations required by the algorithm scales linearly with $\rho$. Furthermore, as $\rho$ increases, condition (\ref{DSPCAgrad-cond}) means that more eigenvalues are required to satisfy to prove convergence, which increases the computational effort.

\begin{figure} [htp]
\begin{tabular*}{\textwidth}{@{\extracolsep{\fill}}cc}
\psfrag{eigs}[b][t]{\small{\% eigs. required}}
\psfrag{time}[t][b]{\small{CPU time (seconds)}}
\includegraphics[width=.50 \textwidth]{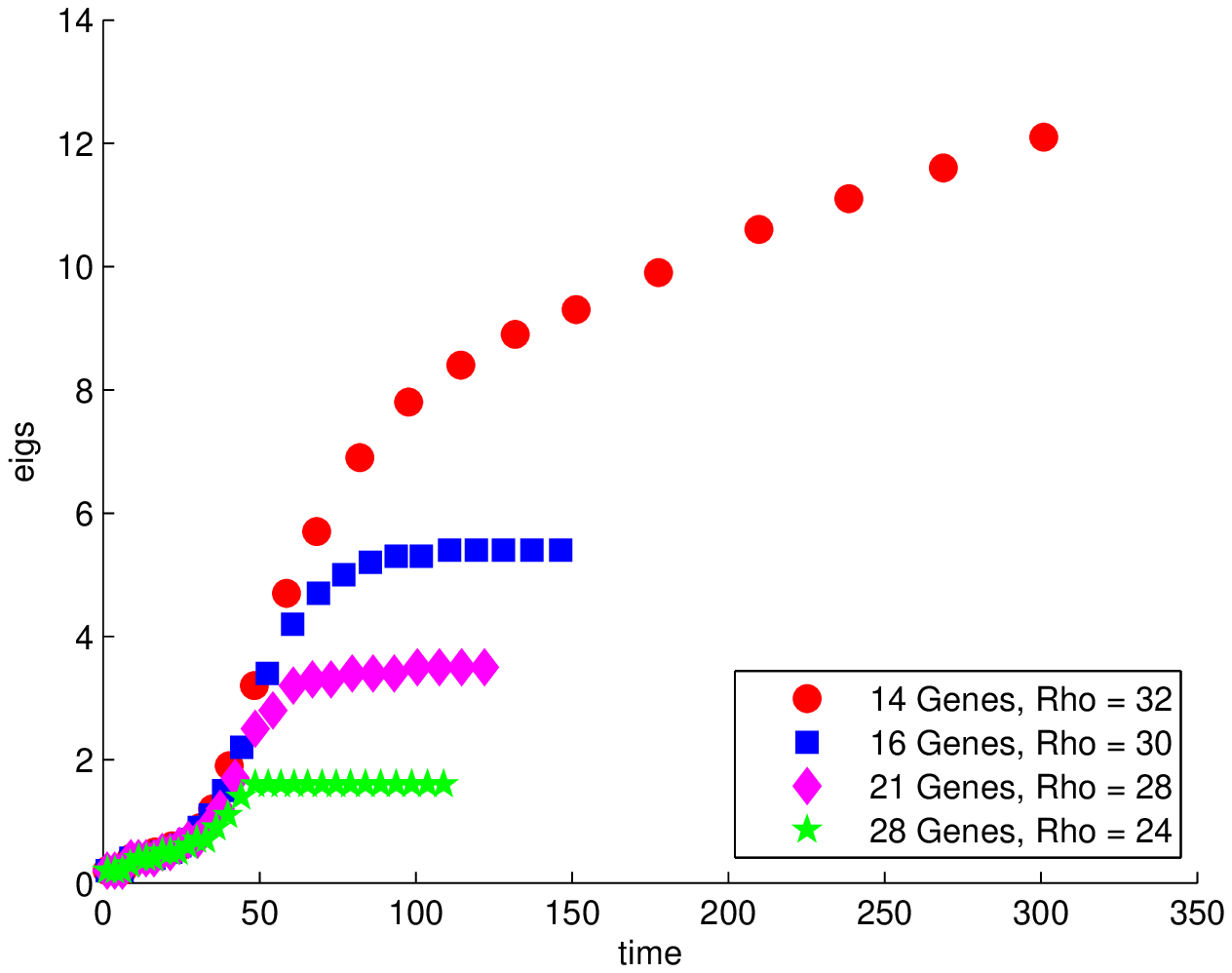} &
\psfrag{eigs}[b][t]{\small{\% eigs. required}} \psfrag{gap}[t][b]{\small{Duality gap}}
\includegraphics[width=.50 \textwidth]{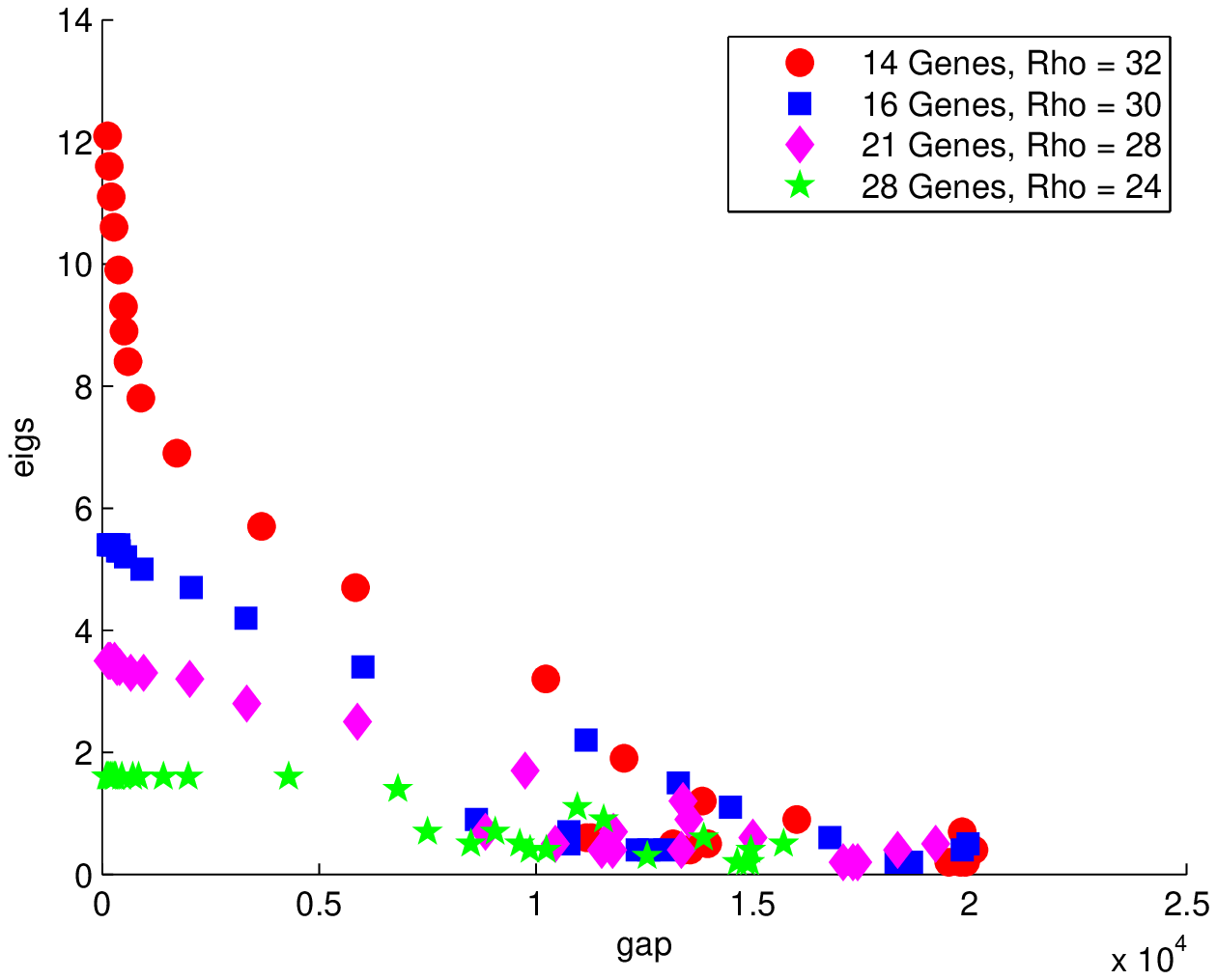}
\end{tabular*}
\caption{Percentage of eigenvalues required versus CPU time (left), and percentage of eigenvalues required versus duality gap (right). The computational cost of an iteration increases with both target precision and target sparsity. }
\label{fig:eigenvalue_coloncancer_figures}
\end{figure}

Figure \ref{fig:eigenvalue_coloncancer_figures} shows how computing time (left) and duality gap (right) vary with the number of eigenvalues required at each iteration (shown as the percentage of total eigenvalues). The four tests are all of dimension 1000 and vary in degrees of sparsity; for a fixed data set at a fixed point in the algorithm, more sparsity (ie. higher $\rho$ and less genes) requires more eigenvalues as noted above. The increase in required eigenvalues is much more pronounced when viewed as the algorithm proceeds, measured by the increasing computing time (left) and decreasing duality gap (right). More eigenvalues are also required as the current iterate gets closer to the optimal solution. Note that, to reduce overhead, the code does not allow the number of required eigenvalues to decrease from one iteration to the next. Figure \ref{fig:eigenvalue_coloncancer_figures} shows as was expected that computational cost of an iteration increases with both target precision and target sparsity.

\subsection{Clustering}
\begin{figure}[htp]
\begin{tabular*}{\textwidth}{@{\extracolsep{\fill}}cc}
\psfrag{fac3}[b][t]{\small{Factor three (500 genes)}} \psfrag{fac2}[t][b]{\small{Factor two (500 genes)}}
\includegraphics[width=.47 \textwidth]{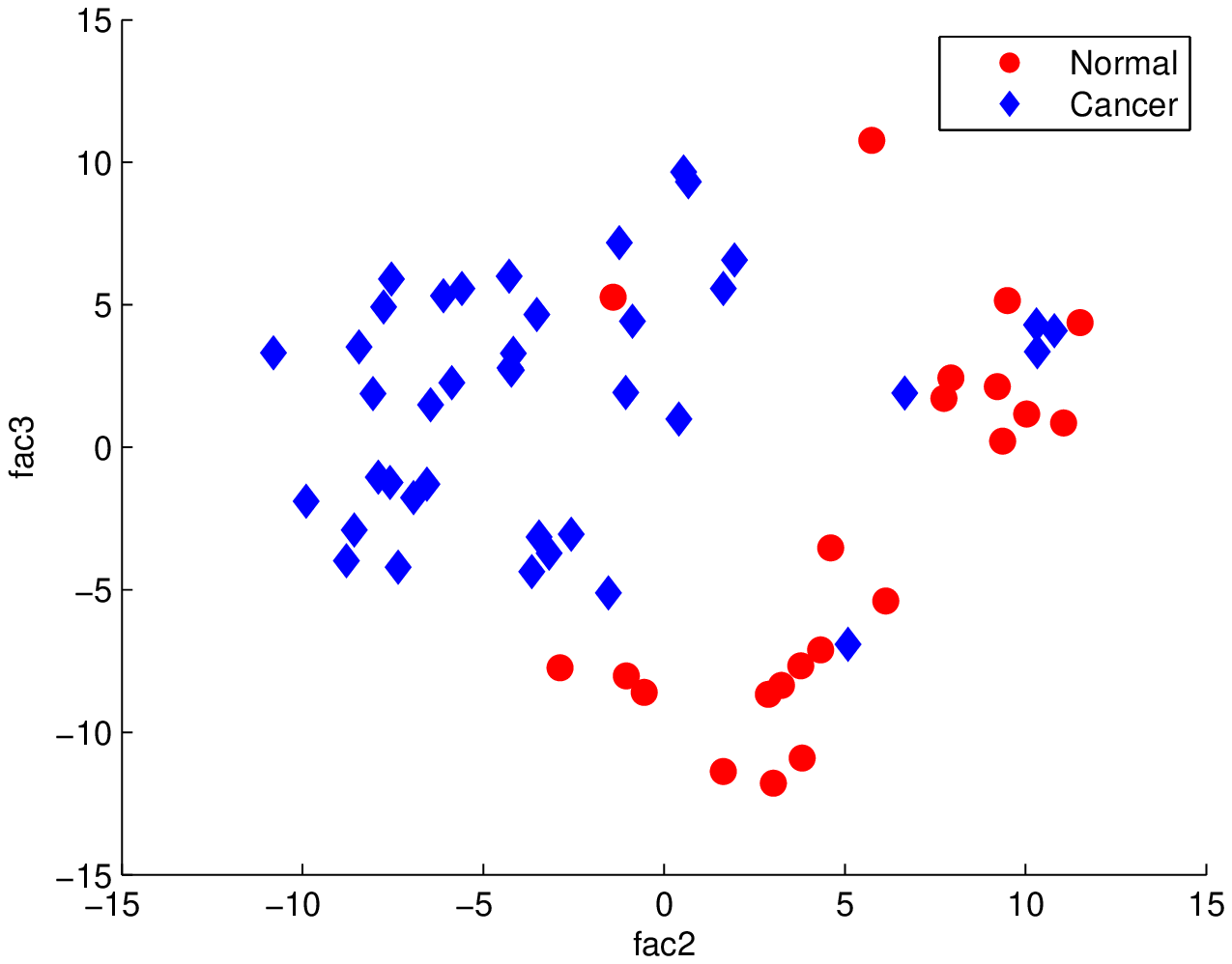} &
\psfrag{fac3}[b][t]{\small{Factor three (1 gene)}} \psfrag{fac2}[t][b]{\small{Factor two (12 genes)}}
\includegraphics[width=.47 \textwidth]{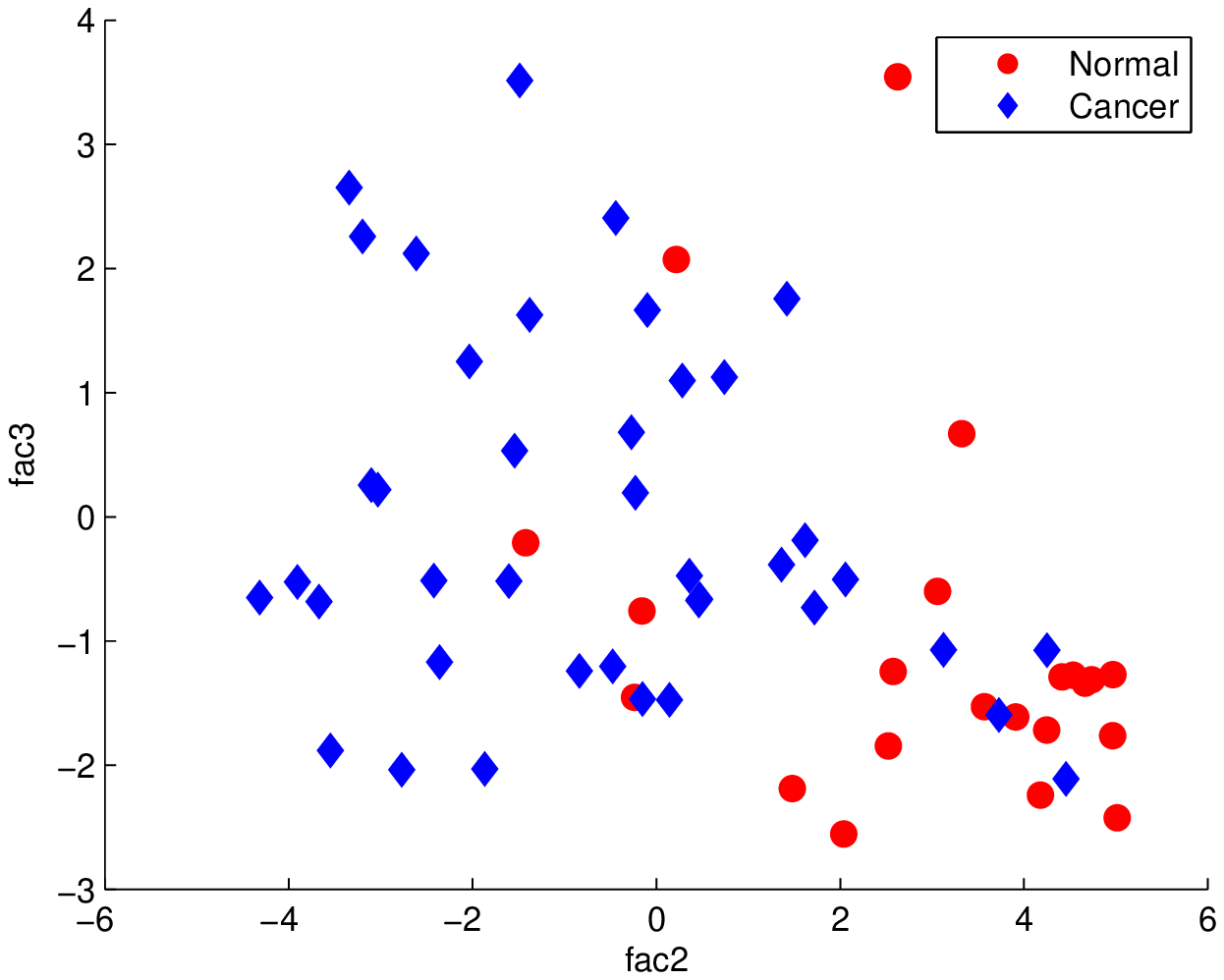} \\
\psfrag{fac2}[b][t]{\small{Factor two (500 genes)}} \psfrag{fac1}[t][b]{\small{Factor one (500 genes)}}
\includegraphics[width=.47 \textwidth]{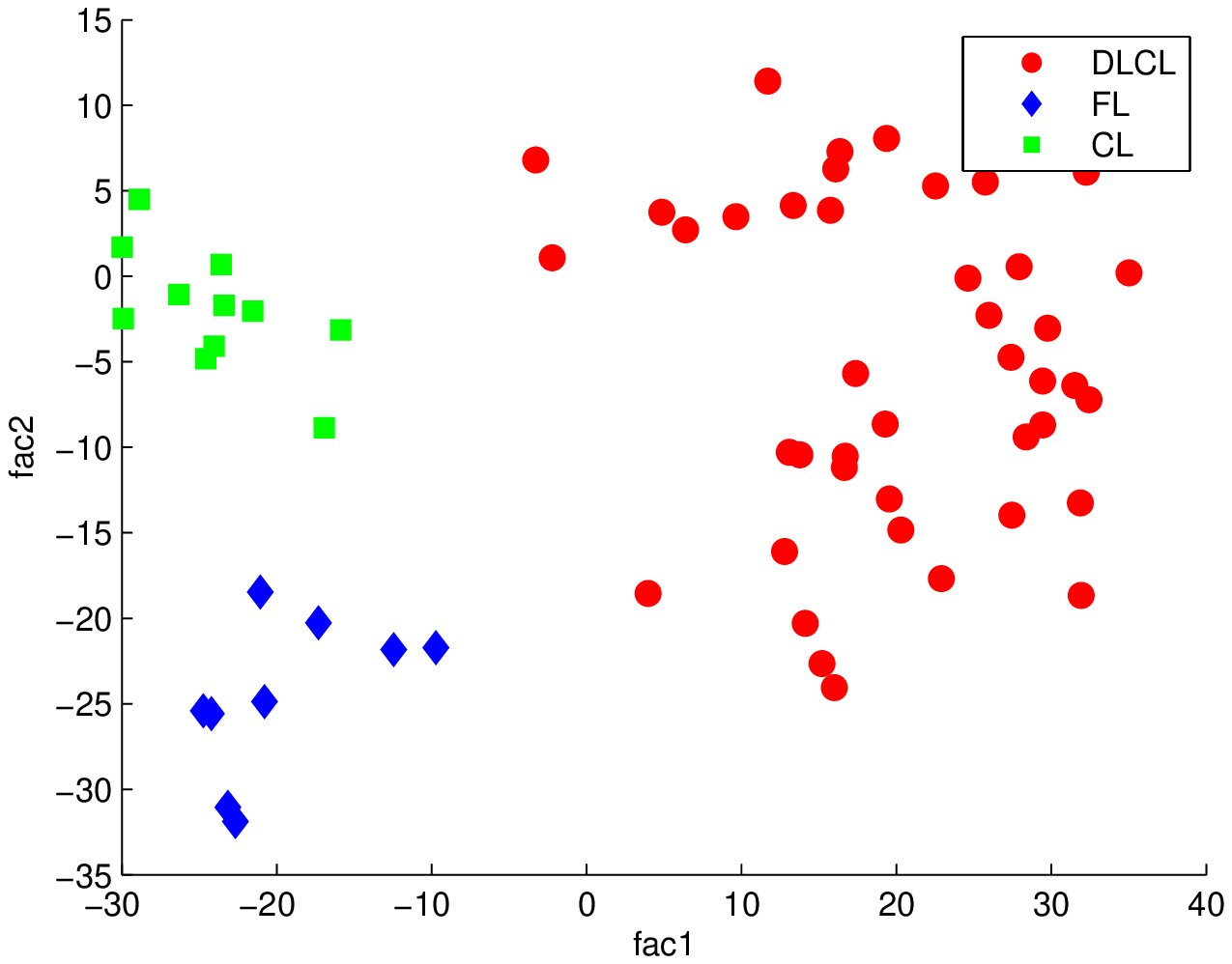} &
\psfrag{fac2}[b][t]{\small{Factor two (53 genes)}} \psfrag{fac1}[t][b]{\small{Factor one (55 genes)}}
\includegraphics[width=.47 \textwidth]{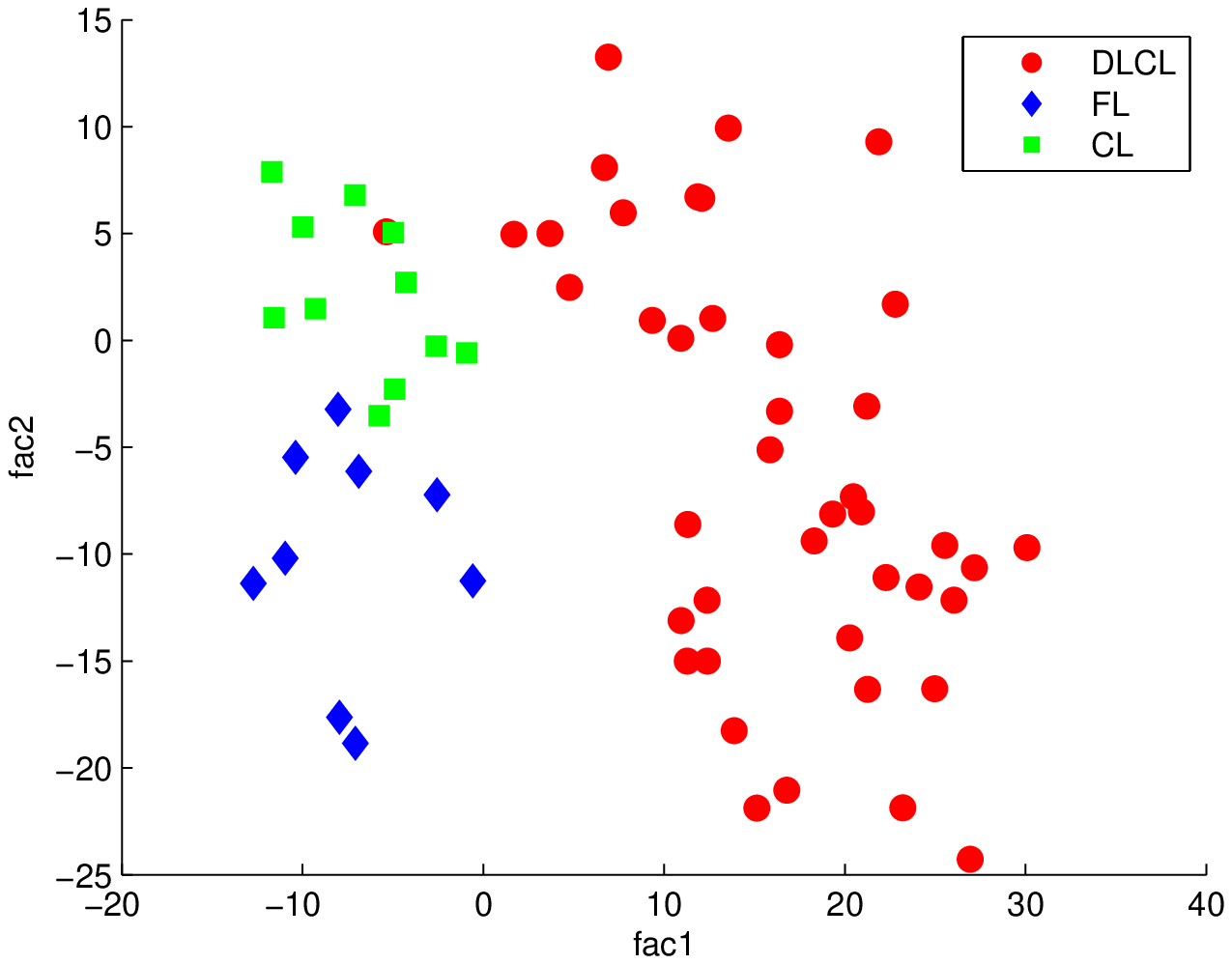}
  \end{tabular*}
\caption{Clustering: The top two graphs display the results on the colon cancer data set using PCA (left) and DSPCA (right).  Normal patients are red circles and cancer patients are blue diamonds.  The bottom two graphs display the results on the lymphoma data set using PCA (left) and DSPCA (right).  For lymphoma, we denote diffuse large B-cell lymphoma as DLCL (red circles), follicular lymphoma as FL (blue diamonds), and chronic lymphocytic leukaemia as CL (green squares).}
\label{fig:PCA_DSPCA_clusters}
\end{figure}

In this section, we compare the clustering (class discovery) performance of sparse PCA to that of standard PCA. We analyze the colon cancer data set of \cite{Alon99} as well as a lymphoma data set from \cite{Aliz00}.  The lymphoma data set consists of 3 classes of lymphoma denoted DLCL (diffuse large B-cell lymphoma), FL (follicular lymphoma) and CL (chronic lymphocytic leukemia).  The top two plots of Figure \ref{fig:PCA_DSPCA_clusters} display the clustering effects of using two factors on the colon cancer data while the bottom two plots of Figure \ref{fig:PCA_DSPCA_clusters} display the results on the lymphoma data. On both data sets, clusters are represented using PCA factors on the left plots and sparse factors from DSPCA on the right plots. For the colon cancer data, the second factor has greater power in predicting the class of each sample, while for the lymphoma data, the first factor classifies DLCL and the second factor differentiates between CL and FL. In these examples, we observe that DSPCA maintains good cluster separation while requiring far fewer genes: a total of 13 instead of 1000 for the colon cancer data, and 108 genes instead of 1000 for the lymphoma data set.

\begin{figure}[htp]
\begin{tabular*}{\textwidth}{@{\extracolsep{\fill}}cc}
\psfrag{rand}[b][t]{\small{Rand index}}
\psfrag{nonzero}[t][b]{\small{Number of nonzero coefficients}}
\includegraphics[width=.47 \textwidth]{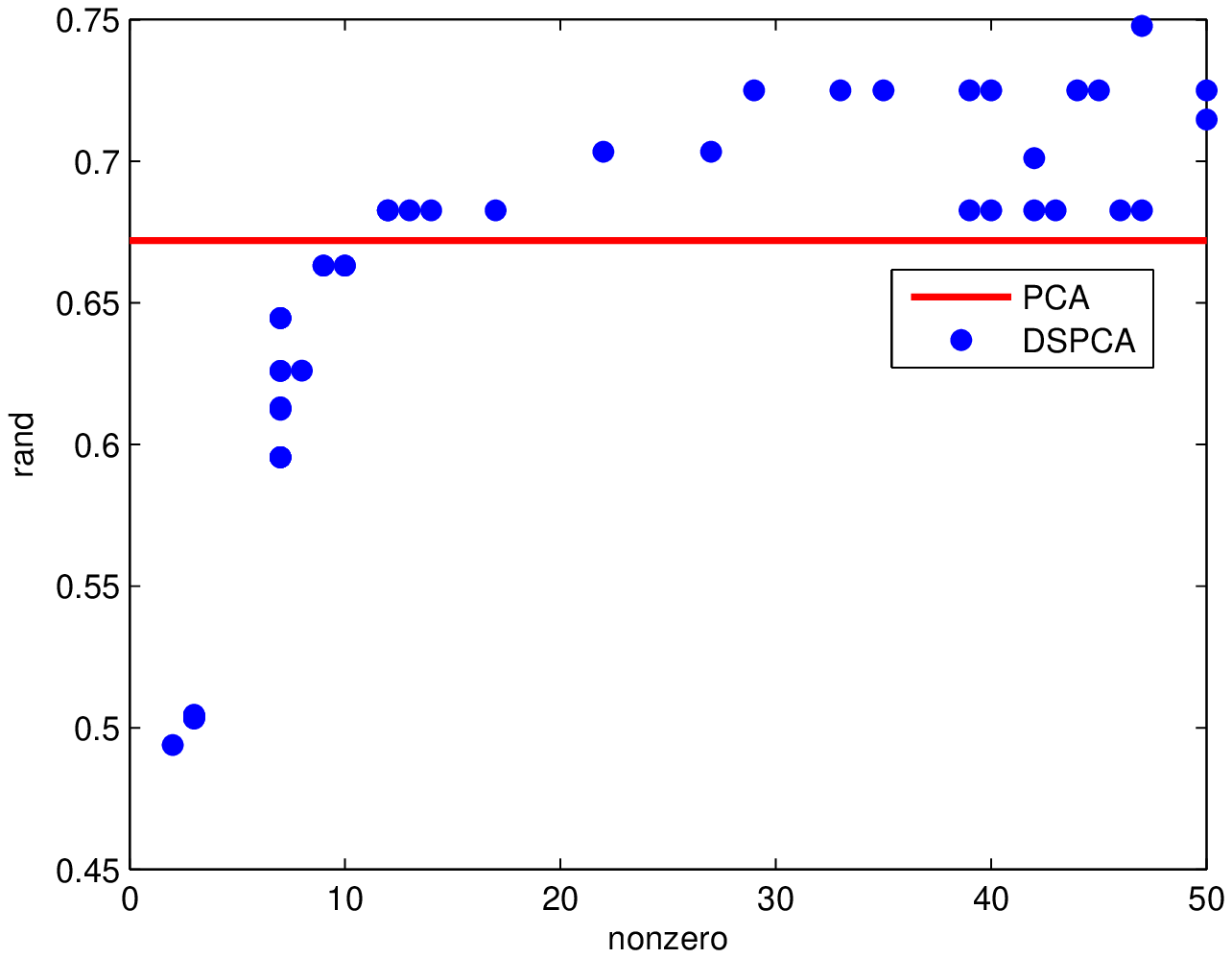} &
\psfrag{rand}[b][t]{\small{Rand index}}
\psfrag{nonzero}[t][b]{\small{Number of nonzero coefficients}}
\includegraphics[width=.47 \textwidth]{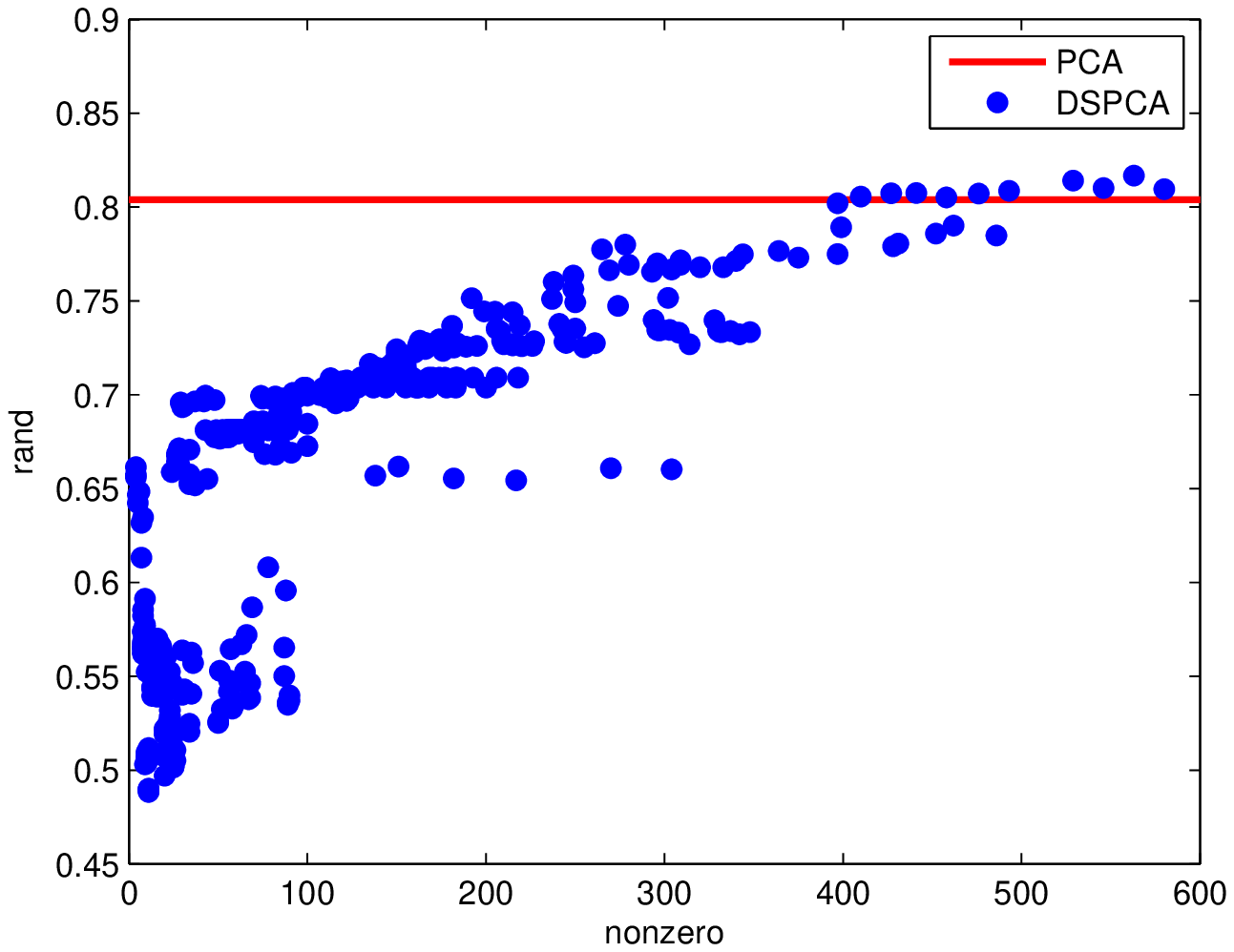}
\end{tabular*}
\caption{Rand index versus sparsity: colon cancer (left) \& lymphoma (right).}
\label{fig:rand}
\end{figure}

While the clustering remains \emph{visually} clear, we now analyze and quantify the quality of the clusters derived from PCA and DSPCA numerically using the Rand index. We first cluster the data (after reducing to two dimensions) using K-means clustering, and then use the Rand index to compare the partitions obtained from PCA and DSPCA to the true partitions.  The Rand index measures the similarity between two partitions $X$ and $Y$ and is computed as the ratio
\[
R(X,Y)=\frac{p+q}{{n \choose 2}}
\]
where $p$ is the number of pairs of elements that are in the same cluster in both partitions $X$ and $Y$ (correct pairs), $q$ is the number of pairs of elements in different clusters in both $X$ and $Y$ (error pairs), and ${n \choose 2}$ is the total number of element pairs. The Rand index for varying levels of sparsity is plotted in Figure \ref{fig:rand}. The Rand index of standard PCA is .654 for colon cancer (.804 for lymphoma) as marked in Figure \ref{fig:rand}.  The Rand index for the DSPCA factors of colon cancer using 13 genes is .669 and is the leftmost point above the PCA line. This shows that clusters derived using sparse factors can achieve equivalent performance to clusters derived from dense factors.  However, DSPCA on the lymphoma data does not achieve a high Rand index with very sparse factors and it takes about 50 genes per factor to get good clusters.

\begin{figure}[htp]
\begin{center}
\psfrag{separation}[b][t]{\small{Separation}}
\psfrag{nonzero}[t][b]{\small{\# of nonzeros}}
\includegraphics[width=.50 \textwidth]{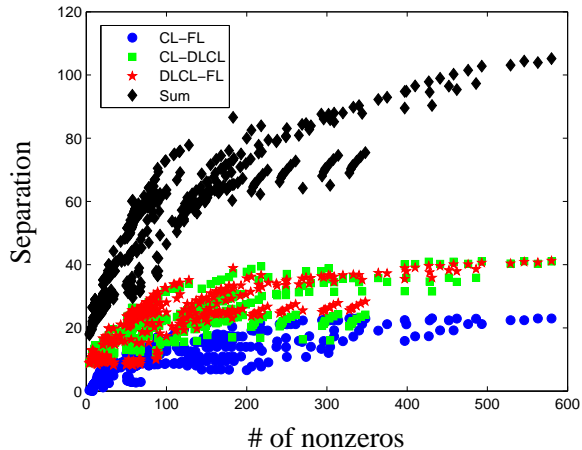}
\caption{Separation versus sparsity for the lymphoma data set.}
\label{fig:separation_lymphoma}
\end{center}
\end{figure}

For lymphoma, we can also look at another measure of cluster validity.  We measure the impact of sparsity on the separation between the true clusters, defined as the distance between the cluster centers. Figure \ref{fig:separation_lymphoma} shows how this separation varies with the sparsity of the factors.  The lymphoma clusters with 108 genes have a separation of 63, after which separation drops sharply. Notice that the separation of CL and FL is very small to begin with and the sharp decrease in separation is mostly due to CL and FL getting closer to DLCL.

\subsection{Feature selection}
Using the sparse factor analysis derived above, our next objective is to track the action of the selected genes. See \cite{Guyo02} for an example illustrating the possible relation of selected genes in this data set to colon cancer using Recursive Feature Elimination with Support Vector Machines (RFE-SVM) for feature selection. Another study in \cite{Huan05a} compares RFE-SVM results on the colon cancer data set to those genes selected by the Rankgene software in \cite{Su03} on the colon cancer and lymphoma data sets.

We implemented RFE-SVM and determined the regularization parameter using cross-validation.  On the colon cancer data, the clustering results show that factor two contains the most predictive power.  Table \ref{tab:rankedgenes_coloncancer} shows 7 genes that appeared in the top 10 most important genes (ranked by magnitude) in the results of DSPCA and RFE-SVM.  For comparison, we include ranks computed using another sparse PCA package: SPCA from \cite{Zou04} with a sparsity level equal to the results of DSPCA (we use the function \emph{spca} for colon cancer and \emph{arrayspca} for lymphoma, both from \cite{Zou05}).  
DSPCA identifies 2 genes from the top 10 genes of RFE-SVM that are not identified by SPCA.  After removing the true first component which explains 58.1\% of the variation, DSPCA, SPCA, and PCA explain 1.4\%, 1.1\%, and 7.2\% of the remaining variation respectively. 

Since the first factor of the lymphoma DSPCA results classifies DLCL, we combine FL and CL into a single group for RFE-SVM and compare the results to factor one of DSPCA.  Table \ref{tab:rankedgenes_lymphoma} shows 5 genes that appeared in the top 15 most important genes of all three feature selection methods.  An additional 2 genes were discovered by DSPCA and SPCA but not by RFE-SVM.  DSPCA, SPCA, and PCA explain 13.5\%, 9.7\%, and 28.5\% of the variation respectively, providing another example where DSPCA maintains more of the clustering effects than SPCA given equal sparsity. Both RFE-SVM and SPCA are faster than DSPCA, but RFE-SVM does not produce factors in the sense of PCA and SPCA is nonconvex and has no convergence guarantees.

\begin{table}[htp]
\begin{center}
\begin{tabular}{|l|l|l|l|l|}
\hline
DSPCA&RFE-SVM&SPCA& GAN & Description \\ \hline
2& 1 & 11& J02854 & \footnotesize{Myosin regulator light chain 2, smooth muscle isoform (human)} \\\hline
4& 3 &9& X86693 & \footnotesize{H.sapiens mRNA for hevin like protein}\\ \hline
5& 6 &1& T92451 & \footnotesize{Tropomyosin, fibroblast and epithelial muscle-type (human)}\\ \hline
7& 4 &5& H43887 & \footnotesize{Complement factor D precursor (H. sapiens)} \\ \hline
8& 9 &NA& H06524 & \footnotesize{Gelsolin precursor, plasma (human)}\\ \hline
9& 2 &NA& M63391 & \footnotesize{Human desmin gene, complete cds} \\ \hline
10& 7 &6&T47377&\footnotesize{S-100P Protein (Human).}\\
\hline
\end{tabular}
\caption{Feature selection on colon cancer data set: Columns 1-3 are respective ranks.  7 genes (out of 2000) were in the top 10 list of genes for both DSPCA (factor two) and RFE-SVM (with C=.005).  5 of the genes were top genes in SPCA.}
\label{tab:rankedgenes_coloncancer}
\end{center}\end{table}

\begin{table}[htp]
\begin{center}
\begin{tabular}{|l|l|l|l|l|}
\hline
DSPCA & RFE-SVM & SPCA & GAN & Description \\ \hline
1& 1 & 1 &GENE1636X & \footnotesize{*Fibronectin 1; Clone=139009} \\ \hline
2& NA & 6 & GENE1637X & \footnotesize{Cyclin D2/KIAK0002=overlaps with middle of KIAK0002}\\
&&&&\footnotesize{cDNA; Clone=359412}\\ \hline
3& 14 &2& GENE1641X & \footnotesize{*Fibronectin 1; Clone=139009}\\ \hline
4& NA &9& GENE1638X & \footnotesize{MMP-2=Matrix metalloproteinase 2=72 kD type IV}\\
&&&&\footnotesize{collagenase precursor=72 kD gelatinase=gelatinase}\\
&&&&\footnotesize{A=TBE-1;Clone=323656}\\ \hline
6& 2 &3& GENE1610X & \footnotesize{*Mig=Humig=chemokine targeting T cells; Clone=8}\\ \hline
10& 3 &4& GENE1648X & \footnotesize{*cathepsin B; Clone=297219} \\ \hline
15& 8 &8& GENE1647X & \footnotesize{*cathepsin B; Clone=261517}\\ \hline
\end{tabular}
\caption{Feature selection on lymphoma data set: Columns 1-3 are respective ranks.  5 genes (out of 4026) were in the top 15 list of genes for both DSPCA (factor one), RFE-SVM (classifying DLCL or not DLCL with C=.001), and SPCA (factor one).  DSPCA and SPCA have an additional 2 top genes not found in the top 15 genes of RFE-SVM.}
\label{tab:rankedgenes_lymphoma}
\end{center}\end{table}

\section{Conclusion}
\label{s:conclusion} 
We showed that efficient approximations of the gradient allowed large scale instances of the sparse PCA relaxation algorithm in \cite{dasp04a} to be solved at a moderate computational cost. This allowed us to apply sparse PCA to clustering and feature selection on two classic gene expression data sets. In both cases, sparse PCA efficiently isolate relevant genes while maintaining most of the original clustering power of PCA.

\section*{Acknowledgements}
We are very grateful to the organizers of the BIRS workshop on Optimization and Engineering Applications where this work was presented. The authors would also like to acknowledge support from grants NSF DMS-0625352, Eurocontrol C20083E/BM/05 and a gift from Google, Inc.

\bibliographystyle{alpha}
\small{\bibliography{DSPCA}}
\end{document}